\documentclass[preprint,10pt,authoryear]{elsarticle}
\usepackage{amssymb}
\usepackage{lineno}
\usepackage[margin=2.5cm]{geometry}
\usepackage[scaled]{helvet}
\usepackage[T1]{fontenc}
\usepackage[hidelinks]{hyperref}
\usepackage{xcolor}
\usepackage{bold-extra}
\usepackage{longtable}
\usepackage{pdflscape}
\usepackage{textcomp}
\usepackage{array}
\usepackage[originalparameters]{ragged2e}
\usepackage{amsfonts}
\usepackage{setspace}
\usepackage{tcolorbox}

\journal{Wiley Interdisciplinary Reviews: Data Mining and Knowledge Discovery on April 18\textsuperscript{th}, 2023}
\begin{document}

\begin{frontmatter}

\title{The role of causality in explainable artificial intelligence}

\author[inst1,inst2,orch1]{Gianluca Carloni}
\author[inst1,inst2]{Andrea Berti}
\author[inst2]{Sara Colantonio}

\affiliation[inst1]{organization={Department of Information Engineering, University of Pisa},
            addressline={Via Caruso 16}, 
            city={Pisa},
            postcode={56122}, 
            country={Italy}}

\affiliation[inst2]{organization={Institute of Information Science and Technologies (ISTI) - National Research Council (CNR)},
            addressline={Via Moruzzi 1}, 
            city={Pisa},
            postcode={56124}, 
            country={Italy}}
\affiliation[orch1]{Corresponding author, ORCiD: 0000-0002-5774-361X}

\begin{abstract}
Causality and eXplainable Artificial Intelligence (XAI) have developed as separate fields in computer science, even though the underlying concepts of causation and explanation share common ancient roots. This is further enforced by the lack of review works jointly covering these two fields. In this paper, we investigate the literature to try to understand how and to what extent causality and XAI are intertwined. More precisely, we seek to uncover what kinds of relationships exist between the two concepts and how one can benefit from them, for instance, in building trust in AI systems. As a result, three main perspectives are identified. In the first one, the lack of causality is seen as one of the major limitations of current AI and XAI approaches, and the "optimal" form of explanations is investigated. The second is a pragmatic perspective and considers XAI as a tool to foster scientific exploration for causal inquiry, via the identification of pursue-worthy experimental manipulations. Finally, the third perspective supports the idea that causality is propaedeutic to XAI in three possible manners: exploiting concepts borrowed from causality to support or improve XAI, utilizing counterfactuals for explainability, and considering accessing a causal model as explaining itself. To complement our analysis, we also provide relevant software solutions used to automate causal tasks.
We believe our work provides a unified view of the two fields of causality and XAI by highlighting potential domain bridges and uncovering possible limitations. 
\end{abstract}


\end{frontmatter}

\textbf{Article Category}: ADVANCED REVIEW.

\textbf{Conflict of Interest}: We declare that we have no conflict of interest.
%
% \newpage
\section{Graphical/Visual Abstract and Caption}
\begin{figure}[h!]
    \centering
    \includegraphics[width=0.599\textwidth]{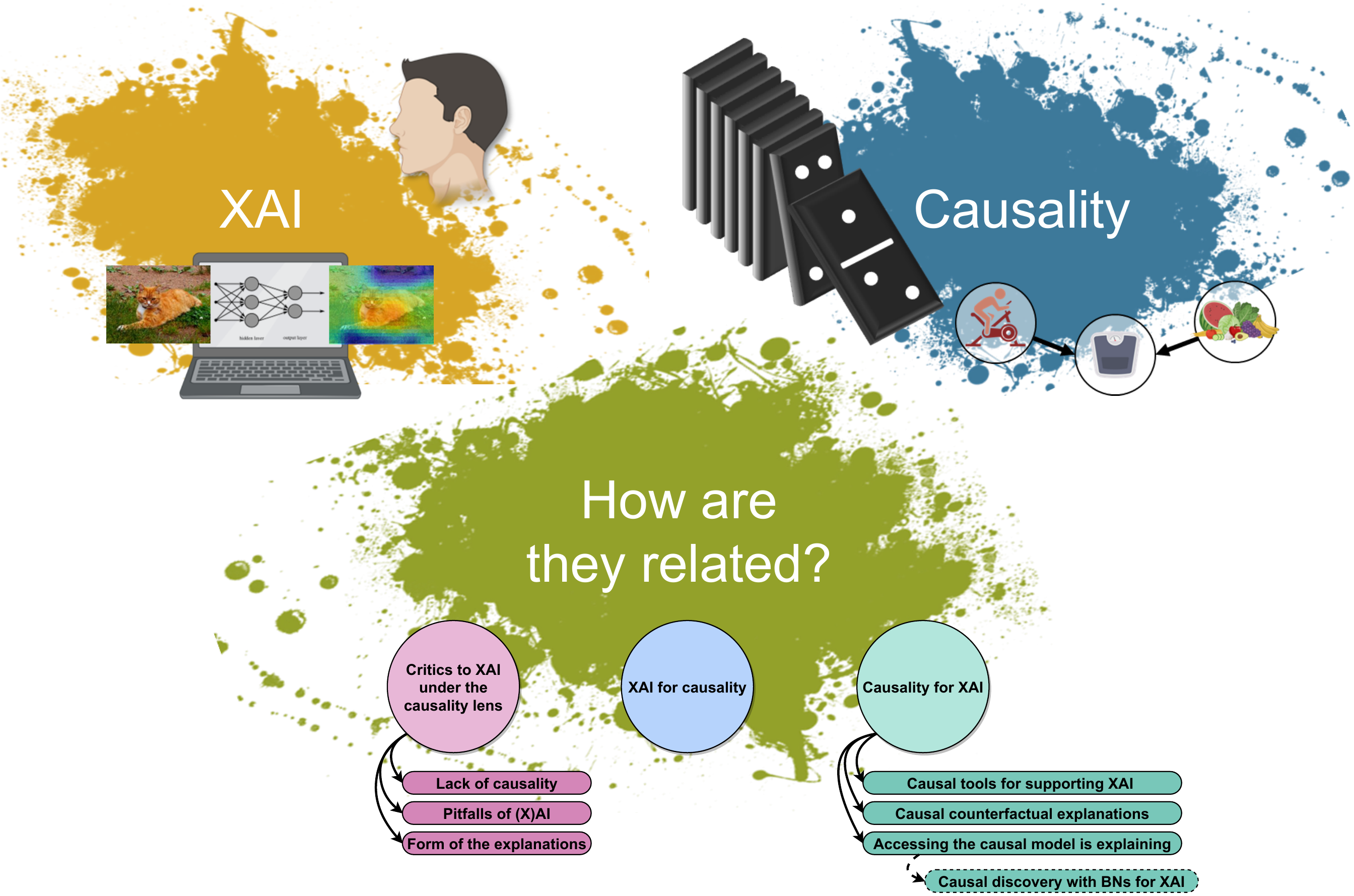}
    \caption{Graphical Abstract: Reviewing the literature to uncover how eXplainable Artificial Intelligence (XAI) and causality are related - the three main perspectives.}
    \label{fig:graphical_abstract}
\end{figure}

% \linenumbers
\section{Introduction}
Causation and explanation are not new concepts, since they have always drawn humans' attention. They are, indeed, highly intertwined since the ancient Greeks and throughout the philosophy of science (Sec.~\ref{sec:intro_rationale_philosophy}). Unfortunately, it seems that these concepts have had a diverse evolution in the field of Artificial Intelligence (AI). Regarding explanations, the eXplainable AI (XAI) research field has been formalized in the past few years to overcome the limitations of conventional black-box machine learning (ML) and deep learning (DL) models (Sec.~\ref{sec:intro_rationale_xai}). Regarding the field of causality (Sec.~\ref{sec:intro_rationale_causality}), some seminal works have been investigating its integration within ML and DL systems \citep{scholkopf2021toward,berrevoets2023causal}. What seems to emerge from the current literature is that there is no clear vision of whether there is a dependent relationship between the two fields.

In this review, we investigate the interdisciplinary literature regarding causality and XAI from both theoretical and methodological viewpoints to try to gain a clearer understanding of this question. Our results show three main perspectives can be identified.
The first way to relate the two fields is to move some critics to XAI under a causal lens, to serve as a watch out. In this regard, a non-negligible subset of publications recognizes causality as a missing component of current XAI research to achieve robust and explainable systems. Other works highlight how the field of XAI (and AI by extension) suffers from certain innate issues, making the problem itself ill-posed. In a similar light, a further branch of works investigates different forms and desiderata of the XAI-produced explanations and their link with the causal theory.
The second perspective tries to relate XAI and causality in a pragmatic way and sees the former as a means to get to the latter. Such works believe XAI has the potential to foster scientific exploration for causal inquiry. Indeed, by means of approaches able to identify pursue-worthy experimental manipulations, XAI may help scientists generate hypotheses about possible causal relationships to be tested.
The third perspective turns the previous one around, claiming that causality is propaedeutic to XAI. Causal tools and metrics are exploited to implement XAI, and specific XAI approaches are brought back to their formal causal definition to improve generalization capabilities. Among the distinctive ideas of this perspective, getting access to the causal model of a system is a way to intrinsically explain the system itself. 

We argue that the third of the perspectives is the one to be preferred to correctly combine the two areas of causality and XAI to advance the research toward reliable systems that are truly useful to humans. Overall, the novelty of our work lies in bridging the XAI-causation gap rigorously, highlighting areas of future development, and exposing limitations.

\section{Rationale and Objective}
\label{sec:intro}

\subsection{Ancient roots}
\label{sec:intro_rationale_philosophy}
The study of causation and explanation can be traced back to the ancient Greek philosophers. Aristotle, for instance, introduced causality as the foundation of explanation and argued that there must be a necessary and sufficient reason for every event \citep{hankinson1998cause}.

As early as the 18th century, the empiricist David Hume formalized causation in terms of sufficient and necessary conditions: an event \textit{c} causes an event \textit{e} if and only if there are event-types \textit{C} and \textit{E} such that \textit{C} is necessary and sufficient for \textit{E}. He, however, remained skeptical about humans' ability to explain and truly know any event. Indeed, he argued that we cannot perceive any necessary connection between cause and effect, but only events occurring in regular succession based on habit \citep{hume2003treatise}.

From the 1950s onward, some others also investigated scientific explanations.
Initially, the “standard model” of explanation was deductive, following the Deductive-Nomological (DN) model by \citet{hempel1948studies}. An outcome was implied logically from universal laws plus initial conditions via deductive inference (e.g., explaining the volume of gas via the ideal gas law and some observations such as pressure).
Regarding Hempel's viewpoint on causality, causal explanations are special cases of DN explanations, but not all laws and explanations are causal.

Later, \citet{salmon1984scientific} developed a model in which good scientific explanations must be statistically relevant to the outcome to get explained. He argued that, in attempting to explain probabilistic phenomena, we seek not merely a high probability but screen for causal influence by removing system components to find ones that alter the probability. Salmon found causality ubiquitous in scientific explanation and was convinced that the time had come to put the “cause” back into “because”. Although remaining vague as to how to attain it, he invited scientists to reconsider the role of causal relations as potentially fundamental constituents of adequate explanations.

\subsection{The need for XAI}
\label{sec:intro_rationale_xai}
Given the rapidly increasing interest in data mining for knowledge discovery, AI is becoming pervasive in our lives, and understanding and trusting its decisions has become imperative. This is further enforced, for instance, by the current guidelines for trustworthy AI by the
European Commission\footnote{\url{https://digital-strategy.ec.europa.eu/en/library/ethics-guidelines-trustworthy-ai}}.
Indeed, opacity in such decisions can lead to reluctance when adopting AI in a product, a decision process, or research. This, therefore, can result in missed opportunities in the use of AI to its fullest potential.
To prevent this scenario, the research field of XAI aims to provide humans with explanations to understand the reasoning behind an AI system and its decision-making process. In other words, the goal of XAI is to enable end-users to understand the underlying explanatory factors of why an AI decision is taken. 
The term XAI was first introduced in \citet{van2004explainable}, but its popularity has spread across the literature only after the DARPA's XAI program \citep{gunning2019darpa}, reaching a certain degree of maturity to date \citep{guidotti2018survey,du2019techniques,carvalho2019machine,rudin2019stop,arrieta2020explainable,molnar2020interpretable_book}. 

XAI systems have been prioritized in different fields, such as healthcare, finance, education, and legal.
In healthcare, XAI has been utilized for medical image analysis, acute critical illness prediction, intraoperative decision support systems, drug discovery, and treatment recommendations \citep{van2022explainable,lauritsen2020explainable,gordon2019explainable,jimenez2020drug}. Regarding finance, popular applications of XAI are credit risk management and prediction, loan underwriting automation, and investment advice \citep{bussmann2021explainable,moscato2021benchmark,sachan2020explainable,yang2021multiple}.
In education, XAI has been applied in automatic essay scoring systems, educational data mining, and adaptive learning systems \citep{kumar2020explainable,alonso2019explainable,khosravi2022explainable}, while digital forensics for law enforcement context represents an example in the legal domain \citep{hall2022explainable}.

Regardless of the application field, XAI is driven by the idea of making the reasoning process of AI transparent and, therefore, AI models more intelligible to humans. Accordingly, when it comes to explaining the logic of an inferential system or a learning algorithm, four aspects can be identified as the main driving motivations for XAI \citep{adadi2018peeking}:
(i) explain to justify (i.e., provide justifications for particular decisions to make sure they are not unfairly yielded by bias),
(ii) explain to control (i.e., understand the system behavior for debugging vulnerabilities and potential flaws),
(iii) explain to improve (i.e, understand the system behavior for enhancing its accuracy and efficiency), and 
(iv) explain to discover (i.e., learn from machines their knowledge on relationships and patterns).

\subsection{A causal perspective}
\label{sec:intro_rationale_causality}
Even though the wide literature on causality spans different interpretations, such as the causal potential theory \citep{xu2018machine} and Wiener-Granger causality \citep{granger1969investigating}, the one by computer scientist Judea Pearl is popularly associated with AI. Pearl identifies some major obstacles still undermining the ability of AI systems in reasoning in a way akin to humans, to be overcome by equipping machines with causal modeling tools \citep{pearl2019seven}. Among those obstacles, is the lack of robustness of AI systems in recognizing or responding to new situations without being specifically programmed (i.e., adaptability), as well as their inability to grasp cause-effect relationships. Instead, those abilities are innate features of human beings, who can communicate with, learn from, and instruct each other since all their brains reason in terms of cause-effect relationships \citep{pearl2018theoretical}.

Pearl argues humans organize their knowledge of the world according to three distinct levels of cognitive ability, which he embodies in distinct rungs of the \textit{Ladder of Causation} \citep{pearl2018book}. As Tab. \ref{tab:ladder_of_causation} shows, the first rung is \textit{Association} and involves passive observation of data. Reasoning on this level could not distinguish the cause from the effect and, although this might come as a surprise to some, Pearl argues that it is where conventional AI approaches to classification or regression stand today. The second rung is \textit{Intervention} and involves not just viewing what exists, but also changing it. However, reasoning on this rung cannot reveal what will happen in an imaginary world where some observed facts are bluntly negated. To this end, we need to climb to the third rung, i.e. \textit{Counterfactuals} (CF). It involves imagination since to answer counterfactual queries one needs to go back in time and change history. For instance, we may wonder whether it was, indeed, \textit{turning the heating system on} that caused a \textit{warm apartment} or, rather, for instance, the outdoor weather.

\begin{table}
\footnotesize 
\caption{The Ladder of Causation by \citet{pearl2018book}.}
\label{tab:ladder_of_causation}
\begin{tabular}{>{\RaggedRight}p{1.75cm}|>{\RaggedRight}p{5cm}|>{\RaggedRight}p{3cm}|>{\RaggedRight}p{5cm}}
\textbf{Level (rung)} & \textbf{Cognitive ability (activity)} & \textbf{Typical questions} & \textbf{Examples}\\
\hline
Association & Seeing, observing (i.e., recognizing recurrent patterns in an environment) & “What if I see \dots?” & "What is the probability that an apartment is warm if I see the heating system being on?"\\
\hline
Intervention & Doing (i.e., predicting the effect(s) of multiple intentional actions on the environment and choosing the best to produce a desired outcome) & "What if I do \dots?" & "What is the probability that the apartment will get warm if I turn on the heating system?".\\
\hline
Counterfactuals & Imagining, reasoning in retrospection, and understanding & "What if I had done \dots?" & "What would have happened to the indoor comfort of the apartment if I had kept the heating system off?".\\
\hline
\end{tabular}
\normalsize
\end{table}

Note that, in a somewhat confusing way, the term "counterfactual" may be encountered also in the XAI literature, where it applies to any instance with an alternative outcome. There, a \textit{counterfactual explanation} (CFE) refers to the smallest change in an input that changes the prediction of an ML classifier \citep{wachter2017counterfactual,mothilal2020explaining}. This concept is quite distinct from the causal meaning of the term. In this regard, as a piece of clarification, we utilize \textit{CFE} and \textit{CF} to address, respectively, the XAI method and the causality concept.

In general, building models that represent causal relationships among variables from observations may be challenging without relying on assumptions that are hard to verify in practice, such as the absence of unmeasured confounding between the variables \citep{robins1999impossibility, greenland2015limitations}. 
Nevertheless, Pearl's work was revolutionary in that it transformed causality from a notion clouded in mystery into a concept with logical foundations and defined semantics. The formalization of causality in mathematical terms within an axiomatic framework allowed the development of automatic computational systems for causal modeling. We refer the reader to \ref{sec:appendix_background} for some notations and terminology regarding Pearl's causality (and related concepts).

\subsection{Objective}
\label{sec:intro_objective}
This review investigates the role(s) of causality in the world of XAI today or, broadly, the relationship between causality and XAI. 
Throughout the paper, we aimed to refer to an interdisciplinary audience, which reflects the use of an accurate (yet not overly zealous) register, leaving the more technical parts (e.g., mathematical notations and supplementary details) to \ref{sec:appendix_background} and \ref{sec:appendix_studyselectionprocess}.

Three main pieces of information led us to believe that those could be complementary fields, and, thus, motivated us to start our investigation.
First, the concepts of causality and explanation have been jointly investigated since ancient times (Sec.~\ref{sec:intro_rationale_philosophy}).
Second, even though they were born separately in the field of AI (\textit{explanation} as XAI and \textit{causation} as Pearl's causality theory), they share a common goal. Indeed, both fields feature human-centricity in AI systems and aim to ensure true usefulness to humans, be it by explaining in a human-comprehensible way what an AI system did, or by designing the system in such a way that it reasons like humans (Sec.~\ref{sec:intro_rationale_xai} --- \ref{sec:intro_rationale_causality}).
Third, another "canary in the coal mine" for us was the presence of the same "counterfactual" term in both fields (Sec.~\ref{sec:intro_rationale_causality}).

To the best of our knowledge, \citet{chou2022counterfactuals} are the only ones investigating a somewhat similar question, albeit with a narrower scope. They systematically review current counterfactual model-agnostic approaches (i.e., CFEs) studying how they could promote \textit{causability}. Causability is a relatively new term representing "the extent to which an explanation of a statement to a human expert achieves a specified level of causal understanding with effectiveness, efficiency, and satisfaction in a specified context of use." \citep{holzinger2019causability}. Since causability differs from causality, this is the first (and major) difference with our study, which covers the wide notion of causality itself. Our work also departs from \citet{chou2022counterfactuals} in that they solely investigate CFE methods, while, in our analysis, we consider the whole corpus of XAI literature, which also includes (but is not limited to) CFEs.

\section{Methods}
\label{sec:methods_eligibility}
This review aims at exploring the literature surrounding the relationship between causality and XAI, from both theoretical and methodological viewpoints.
We conducted our work by adopting a structured process that involved the following: (i) specifying the eligibility criteria; (ii) detailing the information sources; (iii) illustrating the search strategy on specified databases; (iv) describing the selection process; (v) conducting a high-level analysis on the cohort of selected studies; (vi) extracting relevant data and information from studies; and (vii) synthesizing results.

We carried out our search on four popular bibliographic databases,
\textit{Scopus}\footnote{\url{https://www.scopus.com/}}, \textit{IEEE Xplore digital library} (s. IEEE)\footnote{\url{https://ieeexplore.ieee.org/}}, \textit{Web of Science} (s. WoS)\footnote{\url{https://clarivate.com/webofsciencegroup/solutions/web-of-science/}}, and \textit{ACM Guide to Computing Literature} (s. ACM)\footnote{\url{https://dl.acm.org/browse}}, utilizing the following query:\\

\noindent\textsc{(causal*) \textbf{AND} (expla*) \textbf{AND} ("xai" OR "explainable artificial intelligence" OR "explainable ai") \textbf{AND} ("machine learning" OR "ai" OR "artificial intelligence" OR "deep learning")}\\

\noindent Elements within brackets had to be present within at least one of the title, abstract, or keywords of the manuscript. Terms ending with the wildcard “$*$” matched all the terms with the specified common prefix.
Among the obtained publications, we ensured that only peer-reviewed papers from conference proceedings and journals were included. Upon completion of the process of identification, screening, eligibility, and inclusion of articles, $51$ publications formed the basis of our review.
We describe the technical details of the whole study collection process in \ref{sec:appendix_studyselectionprocess}.

In our study, we first performed a high-level analysis of the final cohort of records regarding keywords co-occurrence, then, we extracted information from the publications to answer our research question, and, finally, we collected any cited software solutions in a structured way. 

\subsection{Keywords' co-occurrence analysis}
Regarding the high-level analysis of the final cohort of records, we constructed a bibliometric network of articles' keywords co-occurrence, by utilizing the Java-based application \textit{VOS Viewer}\footnote{\url{https://www.vosviewer.com/}}.
Bibliometric networks are methods to visualize, in the form of graphs, the collective interconnection of specific terms or authors within a corpus of written text. In our setting, we applied such networks to study the paired presence of articles' keywords within a corpus of scientific manuscripts.

\subsection{Research question analysis}
For each of the papers that were included in the review, we identified the most relevant aspects on a conceptual level. According to the research question, we searched for any theoretical viewpoints and comments on the possible ways in which causality and XAI may relate, including formalization frameworks and insights from AI, cognitive, and philosophical perspectives.

Based on the collected information, we performed a topic clustering procedure to organize the literature in related concepts and gain a global view of the field. Selecting cluster topics for a multidisciplinary field as that of causality in the broad field of XAI proved challenging. Topics that are too general would result in an excessively vague and superficial division of papers and therefore be of little use in answering the research question. On the other hand, topics that are too specific would create many quasi-empty clusters, resulting in an improper division, which lacks abstraction capabilities and prevents an overall view of the field. Therefore, we iteratively refined the clusters during a trial-and-error process.

\subsection{Software tools collection}
During the analysis of the full-text manuscripts, we kept track, in a structured collection, of any cited software solutions (e.g., tools, libraries, packages), whenever they were used to automate causal tasks. Specifically, for each one, we analyzed: (i) the URL of the corresponding web-page; (ii) whether the software was commercial or with an open-source license, according to the Open Source Initiative\footnote{\url{https://opensource.org}}; (iii) the name of the company for cases of commercial software; (iv) the eventual release publication that launched the software; (v) whether the frontend consisted in a command line interface (CLI) or a graphical user interface (GUI); and, finally, (vi) the main field of application and purpose.

\section{Results to the keywords' co-occurrence analysis} 
\begin{figure}
    \centering
    \includegraphics[width=0.6\textwidth]{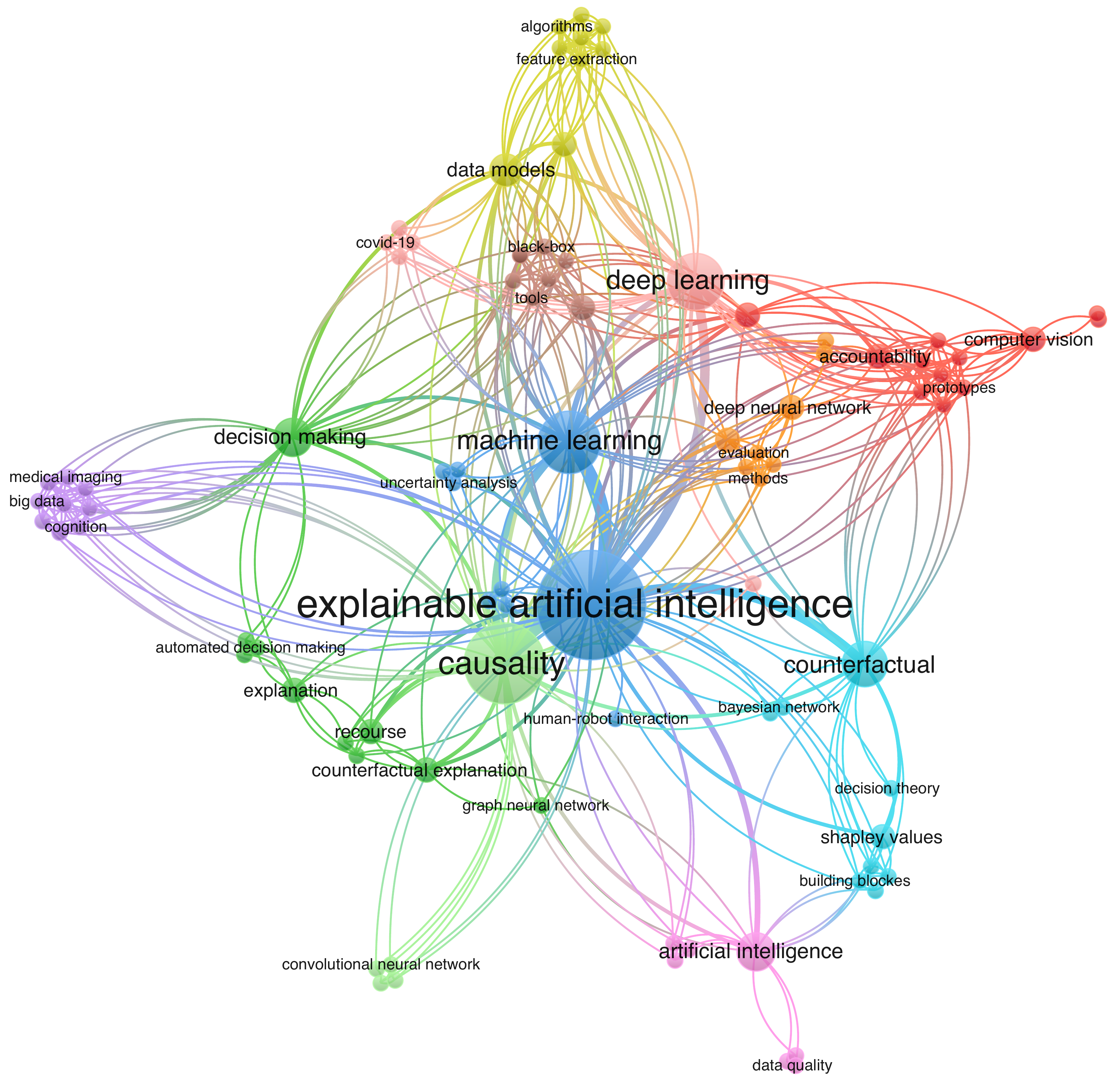}
    \caption{Bibliometric network of papers' keywords for the cohort of publications included in the review.}
    \label{fig:panel_top}
\end{figure}

As a result of the high-level analysis of the final cohort of records, we obtained the bibliometric network shown in Fig.~\ref{fig:panel_top}.
The items (i.e., nodes) of the network represent terms (specifically, articles' keywords); the link (i.e., edge) between two items represents a co-occurrence relation between two keywords; the strength of a link indicates the number of articles in which two keywords occur together; and, finally, the importance of an item is given by the number of links of that keyword with other keywords and by the total strength of the links of that keyword with other keywords. Accordingly, more important keywords are represented by bigger circles in the network visualization, and more prominent links are represented by larger edges between keywords.

\begin{figure}
    \centering
    \includegraphics[width=\textwidth]{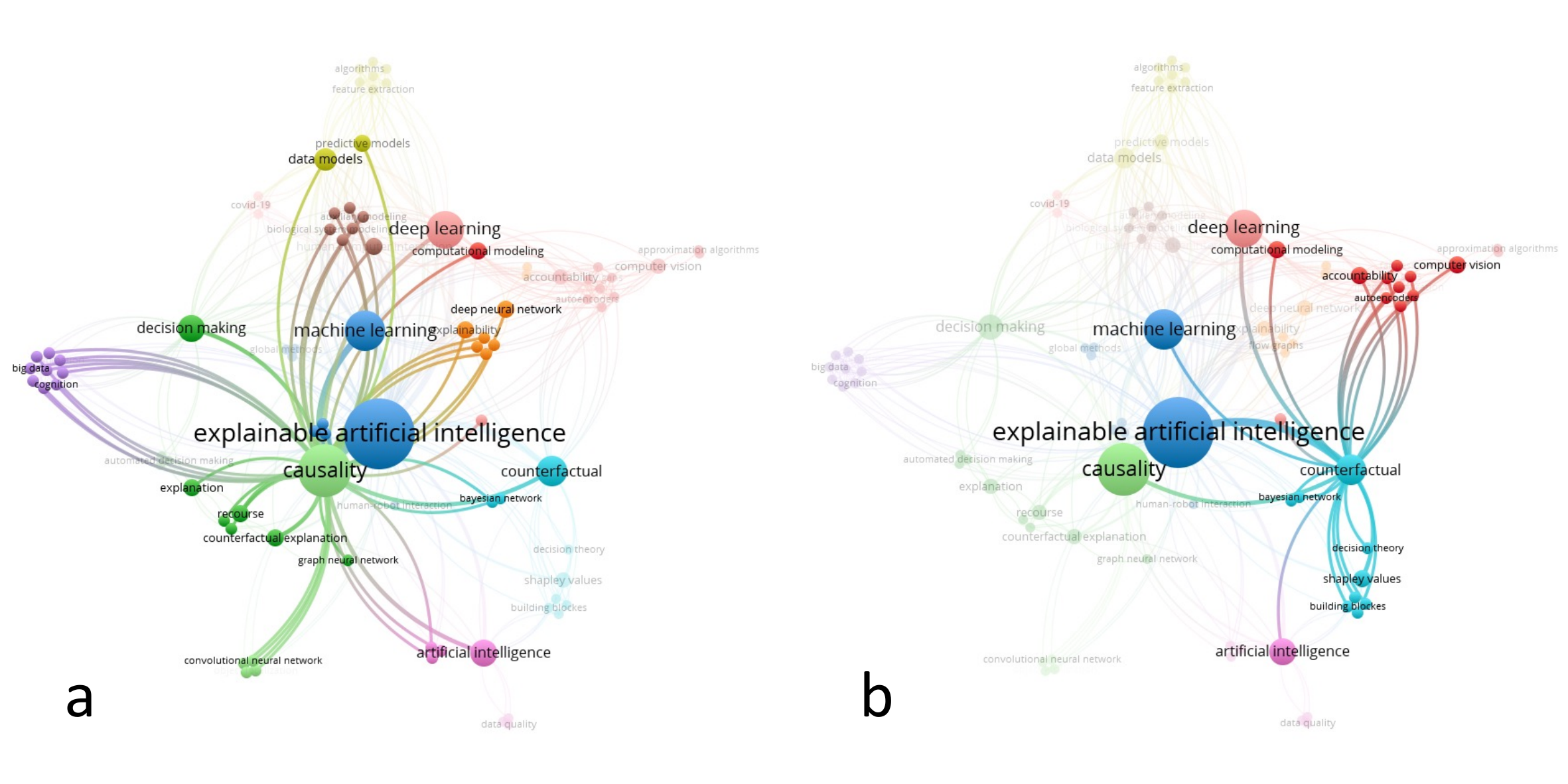}
    \caption{The isolated connections from Fig.~\ref{fig:panel_top} for the terms \textit{causality} (a) and \textit{counterfactual} (b).}
    \label{fig:agglomerato_quadrato}
\end{figure}

This visualization provides insight into how and to what extent the literature relates different research concepts, and it helped us to appreciate the multidisciplinary nature of our research question. Moreover, it is possible to marginalize the scope of specific keywords by identifying the terms to which they relate, as shown in Figs.~\ref{fig:agglomerato_quadrato}a-b for the keywords \textit{causality} and \textit{counterfactual}, respectively.
The relevance and wide scope of the first are justified by the structure of our query, where it was an obligatory search term. Regarding the latter, its scope and relevance represent the central role of the term in both the research fields of causality and XAI. 


\section{Results to the research question analysis}
\begin{figure}[b!]
    \centering
    \includegraphics[width=0.9\textwidth]{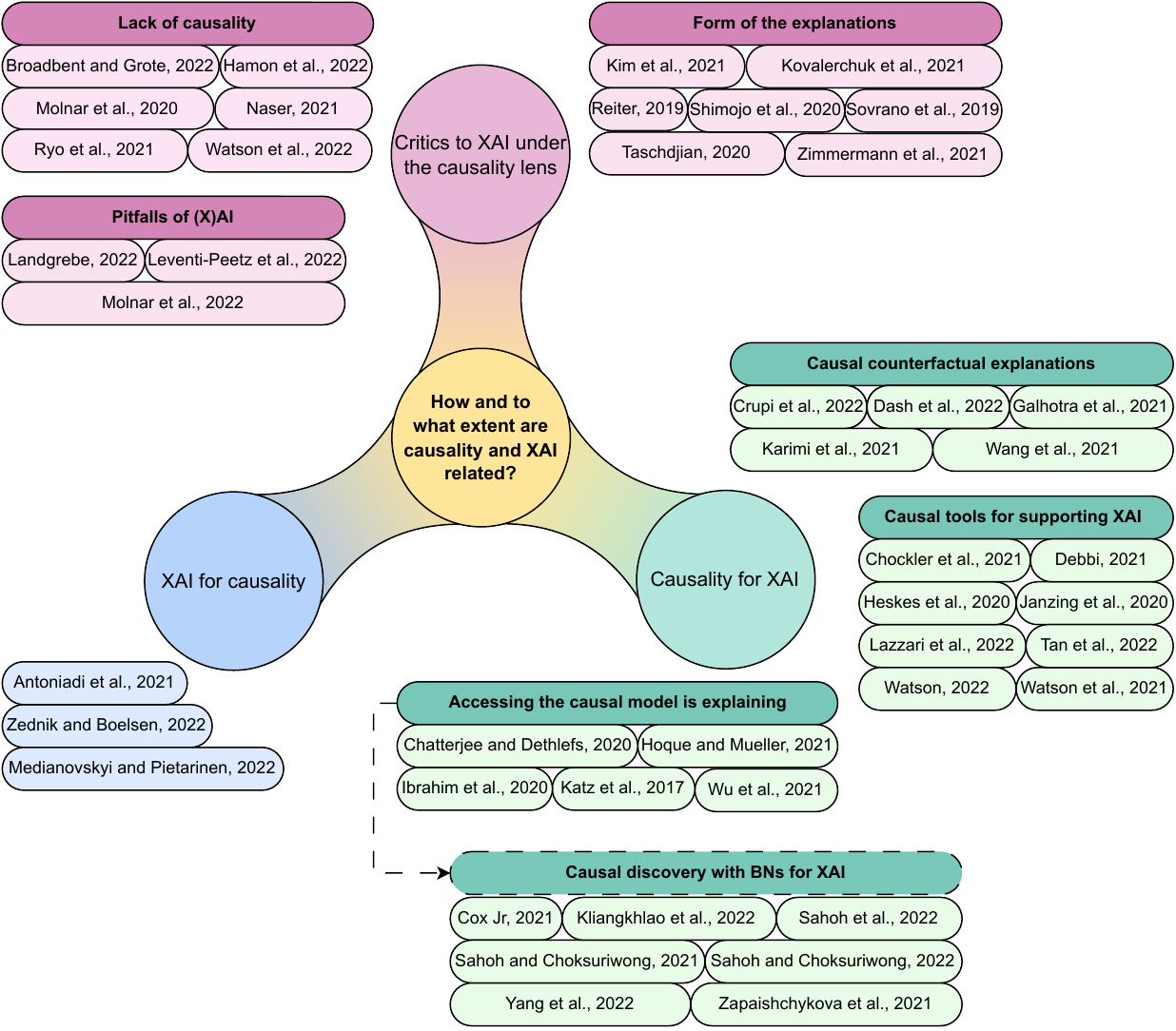}
    \caption{The included studies are classified according to the three main perspectives on how causality and XAI may be related: \textit{Critics to XAI under the causality lens}, \textit{XAI for causality}, and \textit{Causality for XAI}. Next to each of them, are the possible sub-clusters.}
    \label{fig:rq1_clusters}
\end{figure}

This review allowed us to understand how the theory of causality could intertwine with the XAI literature and, specifically, which methodologies and theoretical frameworks could be adopted to approach the bridge between these two fields.
We conceived three main topic clusters of studies, which are presented together with their possible sub-clusters in Fig.~\ref{fig:rq1_clusters}. Specifically, they embody the following perspectives:
\begin{itemize}
    \item \textit{critics to XAI under the causality lens}; 
    
    \item \textit{XAI for causality}; 
    
    \item \textit{causality for XAI}. 
\end{itemize}

This procedure led us to identify which of the three possible perspectives is the preferable one in order to correctly combine the two areas of causality and XAI. We discuss them in Sec.~\ref{sec:rq1_ctx} --- \ref{sec:rq1_cfx}.

\subsection{Critics to XAI under the causality lens}
\label{sec:rq1_ctx}
This first perspective utilizes a causal viewpoint to identify some issues in current XAI. The focus of such papers is either: (i) to point out the inability of XAI to consider causality, (ii) to highlight the profound limitations of current (X)AI both on a methodological and a conceptual level, or (iii) to investigate the forms of the produced explanations.

\subsubsection{Lack of causality}
A fundamental aspect that hinders the value of classical AI models' inference and explainability methods is the lack of a foundation in the theory of causality. Indeed, classical ML and DL predictive models are based on the correlation found among training data instead of true causation. This might be of particular concern in specific fields, such as epidemiology, that have always been grounded in the theory of causation \citep{broadbent2022can}. Moreover, this lack of causality makes models more easily affected by adversarial attacks and less valuable for decision-making \citep{molnar2020interpretable}. Since the parameters and predictions of classical data-driven AI models cannot be interpreted causally, they should not be used to draw causal conclusions.

As \citet{naser2021engineer} points out, meeting specific performance metrics does not necessarily mean that an AI/ML model captures the physics behind a phenomenon. In other words, there is no guarantee that the found correlations map to causal relations between input data and final decisions. For this reason, determining whether such models reflect the true causal structure is crucial \citep{ryo2021explainable}.
This inability of today's ML/DL to grasp causal links reflects also on XAI, constituting a major broad challenge to the ability of AI systems to provide sound explanations.

\citet{hamon2022bridging} stress how this poses serious challenges to the possibility of satisfactory, fair, and transparent explanations. 
Regarding the soundness of the generated explanations, \citet{watson2022agree} demonstrate that they are volatile to changes in model training that are perpendicular to the classification task and model structure. This raises further questions about trust in DL models which just rely on spurious correlations that are made visible via explanation methods.
Since causal explanations cannot be provided for AI yet, explanatory methods are fundamentally limited for the time being. 

\subsubsection{Pitfalls of (X)AI} 
In addition to the weaknesses due to the lack of causality, some works highlight how the fields of AI and XAI may suffer from some innate issues. On a methodological level, \citet{molnar2022general} present a number of pitfalls of local and global model-agnostic interpretation techniques, such as in case of poor model generalization, interactions between features, or unjustified causal interpretations.
At a deeper level, some researchers advocate some concerns about XAI based on its very nature.
For instance, \citet{landgrebe2022certifiable} argues that the human inability to interpret the behavior of deep models in a more objective manner still restricts XAI methods to provide merely a partial, subjective interpretation. Undeniably, deep neural networks solve their classification in a manner that differs completely from the way humans interpret text, language, sounds, and images. For instance, convolutional neural networks (CNNs) use features of the input space to perform their classifications, which are different from those humans use. Not only is it true, but what's more, we do not understand how humans themselves classify texts or images or conduct conversations. Indeed, as of now, human or physical behavior can only be emulated by creating approximations, but approximations cannot be understood any more than complex systems can be.

Under similar considerations, \citet{leventi2022scope} study the scope and sense of explainability in AI systems. In their view, it is impossible or unwise to follow the intention of making every ML system explainable. Indeed, even domain experts cannot always provide explanations for their decisions and, furthermore, on AI systems much higher demands are made than on humans when they have to make decisions.

\subsubsection{Form of the explanations} These works explore different forms, qualities, and desiderata of the explanations produced by XAI methods and their link with causality.
Depending on the application domain, less accurate yet simpler explanations may be preferable to convey a proper understanding of an AI decision. For instance, in Natural Language Generation, a narrative explanation where facts are linked with causal relations is probably a better explanation for narrative-inclined individuals, even though it may not be the most accurate way to describe how the model works \citep{reiter2019natural}. Similarly, in image classification via CNNs, a simpler visualization (e.g., natural dataset examples) may lead to an equal causal understanding of unit activation instead of using complex activation maximization approaches \citep{zimmermann2021well}.

\citet{shimojo2020does} examine what a good explanation is by drawing on psychological evidence regarding two explanatory virtues: (i) the number of causes enforced in an explanation\footnote{This is sometimes referred to as \textit{simplicity} and is conforming with the \textit{Occam's razor} principle, according to which, an event should not be explained by more causes than necessary \citep{jefferys1992ockham}.}, and (ii) the number of effects invoked by cause(s) in an explanation\footnote{This is sometimes referred to as \textit{scope}. Explanations with a broader scope (i.e., correctly predict more events) make humans feel more certain than explanations with a narrower one \citep{johnson2014explanatory}.}. The authors report that, in a user study, the two virtues had independent effects, with a higher impact for the first one.
Similarly, \citet{kim2021multi} discuss several desiderata of XAI systems, among which, they should adjust explanations based on the knowledge of the explainee, to match their background knowledge and expectations. This is further stated by \citet{kovalerchuk2021survey}, who define as "quasi-explanations" those explanations using terms that are foreign to a certain application domain (e.g., medicine, finance, law), such as distances, weights, and hidden layers, and that consequently do make sense only for the data scientists.
\citet{kim2021multi} further states that explanations are considered to be \textit{causal} when they arise from the construction of causal models, serving as the basis for recreating a causal inference chain to (i.e., a “recipe” for reconstructing) a prediction. According to the authors, intelligent systems must be able to provide causal explanations for their actions or decisions when they are critical or difficult to understand.  When a causal explanation answers a "why" question, it can be referred to as a \textit{scientific} explanation. In general, answers to questions such as “How does a personal computer work?” are not considered to be scientific explanations. Such answers are still part of a scientific discipline, but they are descriptive rather than explanatory.

Some other works argue that useful explanations are not only causal explanations but many types of non-causal explanations (e.g., semantic, contrastive, justificatory) may help \citep{sovrano2019difference}.
A pilot user study from \citet{taschdjian2020did} supports this idea revealing that participants preferred causal explanations over the others only when presented in chart form, whilst they resulted as the least favorite choice when in text form.

\subsection{XAI for causality}
\label{sec:rq1_xfc}
Only three papers openly support a pragmatic line of thinking according to which XAI is a basis for causal inquiry. Indeed, such works recognize certain limits of current XAI methods but approach the discussion pragmatically.

\citet{zednik2022scientific} discuss the role of post-hoc analytic techniques from XAI in scientific exploration.
The authors show that XAI techniques, such as CFEs, can serve as a tool for identifying potentially pursue-worthy experimental manipulations within a causal framework and, therefore, for recognizing causal relationships to investigate.
In this regard, the authors remark on an asymmetry between the role of CFEs in \textit{industry} and in \textit{science}. The following two hypothetical scenarios clarify this idea:
\begin{itemize}
    \item \textit{industry}: a bank decides whether to accept or reject a loan application based on an AI agent. A CFE for a rejection case has revealed that doubling the client's income would have led to the acceptance of the loan. Here, the AI agent is not trying to model reality, but it is reality itself. Indeed, a change in the client’s income would actually change the application outcome, meaning that CFEs are \textit{perfect} guides to causal inference.
    \item \textit{science}: an AI agent determines the probability of type-2 diabetes based on patients' features. A CFE for a high-probability case has revealed that losing weight would decrease that probability. Here, the AI agent is trying to model the biological reality of the problem, but still, it remains an approximation. Indeed, it is still possible that losing weight does not actually reduce the probability of type-2 diabetes. That is to say that a change in the model’s behavior does not actually change the way the world works, but at best constitutes a changed representation of how the world could possibly work. In this light, CFEs are \textit{imperfect} guides to causal inference.
\end{itemize}
All in all, it is just because the relevant ML models might not perfectly adhere to reality that the generated XAI explanations only foster scientific \textit{exploration} rather than scientific \textit{explanation}. At most, products of XAI may be thought of as starting points to study potentially causal relationships that have yet to be confirmed.

Similarly, \citet{medianovskyi2022explainable} consider the outputs of the current XAI methods, such as CFEs, to be far from conclusive explanations. Rather, they are initial sketches of possible explanations and invitations to explore further. Those sketches must go through validation processes and experimental procedures before satisfactorily answering the "why" questions, long sought after by XAI.

According to the review by \citet{antoniadi2021current}, XAI can help to shed some light onto causality. Indeed, since causation involves correlation, an explainable ML model could validate the results provided by causality inference techniques. Additionally, XAI can provide a first intuition of (i.e., generate hypotheses about) possible causal relationships that scientists could then test \citep{arrieta2020explainable,lipton2018mythos}.

\subsection{Causality for XAI}
\label{sec:rq1_cfx}
This third perspective is driven by the idea that causality is propaedeutic to XAI. Indeed, these works either: (i) exploit causality-based concepts to support XAI, (ii) restore the causal foundation of CFEs, or (iii) argue that accessing the causal model of a system is intrinsically explaining the system itself.

\subsubsection{Causal tools for supporting XAI}
Such papers interpret the role of causality in XAI in the sense that some causal concepts, such as structural causal model (SCM) and \textit{do}-operator (\ref{sec:appendix_background_scm}) and causal metrics, may bring useful tools for explainability and for finding the causes of AI predictions. 
Regarding the use of \textbf{Structural causal models} to foster XAI, \citet{reimers2020determining} reduce DL to a basic level and frame the constitutional structure of a CNN model into an SCM. In this setting, the random variables represent, for instance, the network's weights and the final prediction, while the functions linking the variables are the \textit{training function} (from labeled images to the network's weights), and the \textit{inference function} (from unlabeled images and weights to the prediction).
By doing so, the authors aim to establish whether a feature is relevant to a CNN prediction by leveraging causal inference and Reichenbach's Common Cause Principle\footnote{According to \citet{reichenbach1956direction}, if two variables A and B are dependent, then there exists a variable C that causes A and B. In particular, C can be identical to A or B meaning that A causes B or B causes A.}.

\citet{lazzari2022predicting}, in order to predict employee turnover, utilize the concept of SCM to revisit and equip the Partial Dependence Plot (PDP)\footnote{A visual tool introduced by \citet{friedman2001greedy}, commonly used for model-agnostic XAI, that shows the marginal effect of one feature on the predicted outcome of a system.} method with causal inference properties. Their SCM-based PDP can now go beyond correlation-based analyses and reason about causal interventions, allowing one to test causal claims around factors. This, in turn, provides an intuitive visual tool for interpreting the results and achieving the explainability of automatic decisions.

Regarding \textbf{\textit{do}-operator}, some authors employ this concept to bring the theory of Shapley values a step further. A fundamental component of Shapley values is to evaluate the reference distribution of dropped (i.e., 'out-of-coalition') features, which has implications on how Shapley values are estimated since this helps define the value function. Based on this distribution, the following variants of Shapley values exist \citep{watson2022rational,heskes2020causal}: \textit{marginal} Shapley values (they ignore relations among features and are used to discover the model's decision boundary), \textit{conditional} Shapley values (they consider feature dependencies and condition by observation), and \textit{interventional} Shapley values. The latter was introduced by \citet{janzing2020feature} who replaced conventional \textit{conditioning by observation} with \textit{conditioning by intervention} (\textit{do}-operator).

Extending this concept, \citet{heskes2020causal} introduce \textit{causal} Shapley values by explicitly considering the causal relationships between the data in the real world to enhance the explanations. Using the interventional distribution is optimal when, with access to the underlying SCM, one seeks explanations for causal data-generating processes. These methods are required when seeking to use XAI for discovery and/or planning, as they seem to provide sensible, human-like explanations that incorporate causal relationships in the real world. 

Finally, some other works borrow \textbf{metrics from the causal theory} to aid XAI, and, specifically, Probability of Necessity (PN) and Probability of Sufficiency (PS) from \citet{glymour2016causal} and the metric of \textit{responsibility} from  \citet{chockler2004responsibility}.
Regarding PN and PS, two works investigate their implications for XAI. Indeed, such probabilities, often addressed as "probabilities of causation", play a major role in all "attribution" questions.
\citet{watson2021local} formalize the relationship between existing XAI methods and the probabilities of causation. For instance, they highlight the role of PN and PS in feature attribution methods and CFEs. 
Regarding the former, the authors reformulate the theory of Shapley values in their framework and show how the value function (i.e., the payoff associated with a feature subset) precisely corresponds to the PS of a factor.
Regarding the latter, the authors rewrite the CFE optimization problem with an objective based on the PS of the factor with respect to the opposite of the outcome.
Moreover, \citet{tan2022learning} borrow PN and PS and adapt them to evaluate the necessity and sufficiency of the explanations extracted for a graph neural network (GNN). This makes it possible to conduct a quantitative evaluation of GNN explanations even without ground-truth explanations for real-world graph datasets.

On the other hand, regarding the metric of responsibility, \citet{chockler2021explanations} propose \textsc{DC-Causal}, a greedy, compositional, perturbation-based approach to computing explanations for image classification. It leverages causal reasoning in its feature masking phase with the goal of finding causes in input images by causally ranking parts of the input image (i.e., superpixels) according to their responsibility for the classification.
In addition to responsibility, \citet{debbi2021causal} borrows from \citet{chockler2004responsibility} the concept of blame to compute visual explanations for CNN decisions. The author abstracts the CNN model into a causal model by virtue of similarity in a hierarchical structure, and filters are considered as actual causes for a decision. First, each filter is assigned a degree of responsibility (i.e., weight) as a measure of its importance to the related class. Then, the responsibilities of these filters are projected back to compute the blame for each region in the input image. The regions with highest blame are returned then as the most important explanations.

\begin{tcolorbox}[width=\textwidth,colback=red!5!white,colframe=red!75!black, before upper={\parindent15pt}]    
   % \blindtext[1]
   PN is the probability that the garden would not have got wet had the sprinkler not been activated ($Y_0=0$), given that, in fact, the garden did get wet ($Y=1$) and the sprinkler was activated ($X=1$). Mathematically, this becomes: $PN = P(Y_0=0|X=1,Y=1)$.
In other words, this probability quantifies to what extent activating the sprinkler is necessary to get the garden wet, and consequently if other factors (e.g., rain) may have caused the wet garden.

PS is the probability that the garden would have got wet had the sprinkler been activated ($Y_1=1$), given that the sprinkler had not in fact been activated ($X=0$), and the garden did not get wet ($Y=0$). Mathematically, this becomes: $PS = P(Y_1=1|X=0,Y=0)$.
In other words, this probability quantifies to what extent activating the sprinkler is sufficient to wet the garden, and consequently, if there may exist scenarios (e.g., hardware malfunctioning) where activating the sprinkler does not wet the garden.

Responsibility is a quantification of causality, attributing to each actual cause its degree of responsibility $\frac{1}{1+k}$, which is based on the size $k$ of the smallest contingency feature set required to obtain a change in the prediction (i.e., creating a counterfactual dependence).
The degree of responsibility is always between $0$, for variables that have no causal influence on the outcome ($k \rightarrow\infty$), and $1$, for counterfactual causes ($k=1$). Responsibility extends the actual causality framework of \citet{halpern2005causes}
.

\end{tcolorbox} 

\subsubsection{Causal counterfactual explanations}
As noted in Sec.~\ref{sec:intro_rationale_causality}, the \textit{counterfactual} concept seems to belong both to the XAI literature and to the causality literature. 
Some authors remark on how CFEs and CF are two separate concepts \citep{crupi2022leveraging} and, strictly speaking, some would not even call the former \textit{counterfactuals}, precisely to contrast the causal perspective \citep{dash2022evaluating}.
Interestingly, however, these two seemingly separate concepts may be bridged in what we could name structural causal explanations. Indeed, the papers in this sub-cluster present methods for generating CF based on their formal causal definition, restoring the causal underpinning to CFEs by using the concept of SCM and Pearl's CF three-step "recipe" (\ref{sec:appendix_background_scm}).

In their quest to explain an image classifier's output and its fairness using counterfactual reasoning, \citet{dash2022evaluating} propose \textsc{ImageCFGen}, a system that combines knowledge from an SCM over image attributes and uses an inference mechanism in a generative adversarial network-like framework to generate counterfactual images.
The proposed architecture directly maps to Pearl's three steps: (i) for \textit{abduction}, an encoder infers the latent vector of an input image coupled with its attributes; (ii) for \textit{action}, a subset of desired attributes is changed and, accordingly, the values of their descendants in the SCM are updated; (iii) for \textit{prediction}, a generator takes the latent vector together with the modified set of attributes and produces a counterfactual image.
A subset of work focuses on a specific aim of the XAI research tightly bound with counterfactual reasoning, i.e., \textit{recourse}. Recourse can be seen as the act of recommending a set of feasible actions to assist an individual to achieve a desired outcome. 
\citet{karimi2021algorithmic} argue that the conventional, non-causal CFEs are unable to convey a relevant recourse to the end-user of AI algorithms since they help merely understand rather than act (i.e., inform an individual to where they need to get, but not how to get there). Shifting from explanation to \textit{minimal intervention}, the authors leverage causal reasoning (i.e., tools of SCMs and structural interventions) to incorporate knowledge of the causal relationships governing the world in which actions will be performed. This way, the authors are able to compute what they refer to as \textit{structural CF} by performing the \textit{abduction}-\textit{action}-\textit{prediction} steps and provide \textit{algorithmic recourse}.
\citet{galhotra2021explaining} introduce \textsc{Lewis}, a principled causality-based approach for explaining black-box decision-making systems. They propose to achieve \textit{counterfactual recourse} by solving an optimization problem that searches for minimal interventions on a pre-specified set of actionable variables that have a high probability of producing the algorithm's desired future outcome. Notably, the authors propose a GUI that implements \textsc{Lewis}, of which they show a demo in \citet{wang2021demonstration}.
\citet{crupi2022leveraging} also contribute to the recourse objective by proposing \textsc{Ceils}, a new post-hoc method to generate causality-grounded CFEs and recommendations. It involves the creation of an SCM in the latent space, the generation of causality-grounded CFEs, and their translation to the original feature space.

\subsubsection{Accessing the causal model is explaining}
\label{par:res_rq1_cfx_accessing}
Part of the work relates to the common thought that accessing the causal model of a system intrinsically explains the system itself. Under this view, two fundamental observations are supported:
\begin{itemize}
    \item when a model is built on a causal structure, it is inherently an interpretable model;
    \item making the inner workings of a causal model directly observable, such as through a directed acyclic graph (DAG) (\ref{sec:appendix_background_dag}), makes the model inherently interpretable.
\end{itemize}

Much of the causality theory focuses on explaining observed events, that is, inferring causes from effects. According to its retrospective attribution, causality lies at the heart of explanation-based social constructs such as explainability and, therefore, causal reasoning is an important component of XAI \citep{wu2021methods}.

\citet{ibrahim2020actual} try to fill the lack in the causality literature of automatic and explicit operationalizations to enable explanations. The authors propose an extensible, open-source, interactive tool (Actual Causality Canvas) able to implement three main activities of causality (causal modeling, context setting, and reasoning) in a unifying framework. According to the authors, what Canvas can provide, through answers to causal queries, largely overlaps with the ultimate goal of XAI, which is providing the end-user with explanations of why particular factors occurred.
\citet{hoque2021outcome} propose Outcome Explorer, an interactive framework guided by causality, that allows expert and non-expert users to select a dataset, choose a causal discovery (CD) algorithm for structure discovery (\ref{sec:appendix_background_bn}), generate (and eventually refine) a causal diagram, and interpret it by setting values to the input features to observe the changes in the outcome.
\citet{katz2017autonomous} propose an XAI system that encodes the causal relationships between actions, intentions, and goals from an autonomous system and explains them to a human end-user with a cause-effect reasoning mechanism (i.e., causal chains).  
\citet{chatterjee2020temporal} exploit the representational power of CNNs with attention, to discover causal relationships across multiple features from observed time-series and historical error logs. The authors believe causal reasoning can enhance the reliability of decision support systems making them more transparent and interpretable.

A subset of publications sees CD as the most appropriate way of operationalizing the idea that accessing the causal model of a system intrinsically explains the system itself. In this regard, all of them utilize \textbf{Bayesian networks (BNs)} (\ref{sec:appendix_background_bn}) as the methodological tool.
Since establishing unique directions for edges based on passive evidence alone may be challenging, knowledge-based constraints can help orient arrows to reflect causal interpretations \citep{cox2021information}. In line with this, some works perform CD with BNs in a mixed approach: on the one hand, they leverage knowledge from domain-experts to outline the causal structure of the system (i.e., finding nodes and related edges); on the other hand, they fit the model parameters on observed, real-world data.

\citet{sahoh2022proof} propose a new system to support emergency management (e.g., terrorist events) based on the Deep Event Understanding perspective, introduced in an earlier work of theirs \citep{sahoh2021beyond}. Deep Event Understanding aims to model expert knowledge based on the human learning process and offers explanation abilities that mimic human reasoning. Their model utilizes BNs based on social sensors as an observational resource (i.e., text data from Twitter), with prior knowledge from experts to infer and interpret new information. Their approach helps in recognition of an emergency event and in the uncovering of its possible causes, contributing to the explanation of “why” questions for decision-making.

\citet{sahoh2022causal} propose discovering cause-effect ML models for indoor thermal comfort in Internet of Things (IoT) applications. They employ five different CD algorithms and show how these may converge to the ground-truth SCM of the problem variables obtained from domain experts.
\citet{kliangkhlao2022design} introduce a BN model for agricultural supply chain applications, initially constructed from causal assumptions from expert qualitative knowledge, which conventional ML cannot reasonably conceive. Therefore, a data-driven approach using observational evidence is employed to encode these causal assumptions into quantitative knowledge (i.e., parameter fitting). The authors report their system constitutes a framework that is able to provide reasonable explanations of events for decision-makers.

In \citet{zapaishchykova2021interpretable} the authors leverage the respective strengths of DL for feature extraction and BNs for causal inference, achieving an automatic and interpretable system for grading pelvic fractures from CT images. The BN model is constructed upon variables extracted with the neural network, together with a variable from the clinical practice (i.e., patient age). By doing so, the authors believe that the framework provides a transparent inference pipeline supplying fracture location and type, by establishing causal relationships between trauma classification and fracture presence.

\citet{yang2022interpretable} propose a new process monitoring scheme based on BNs to explain (diagnose) a detected fault and promote decision-making. Their system allows the identification of the root cause (i.e., labeling the abnormal variables)  so that the result of the analysis can be linked to the repairing action, reducing the investigation time. Among one of their use cases, the authors fit a BN model on observed, real-world data for manufacturing fault events. During this CD process, they employ a blacklist obtained from domain experts to exclude causally-unfeasible relationships.

\section{Results of software tools collection}
We hereby present a summary of the main data mining software tools collected within the cohort of papers. Table~\ref{tab:software_tools_qualities} comprises tools for
performing CD with BNs (i.e., PySMILE\footnote{\url{https://www.bayesfusion.com/smile/}}, CausalNex\footnote{\url{https://causalnex.readthedocs.io/en/latest}}, bnlearn\footnote{\url{https://www.bnlearn.com}}, CompareCausalNetworks\footnote{\url{https://cran.r-project.org/web/packages/CompareCausalNetworks/}}, CaMML\footnote{\url{https://bayesian-intelligence.com/software/}}, Python Causal Discovery Toolbox\footnote{\url{https://fentechsolutions.github.io/CausalDiscoveryToolbox/html/index.html}}, and Tetrad\footnote{\url{https://htmlpreview.github.io/?https:///github.com/cmu-phil/tetrad/blob/development/docs/manual/index.html}}),
creating and analysing SCMs (i.e., IBM\textsuperscript{\tiny\textregistered} SPSS\textsuperscript{\tiny\textregistered} Amos\footnote{\url{https://www.ibm.com/products/structural-equation-modeling-sem}}, lavaan\footnote{\url{https://cran.r-project.org/web/packages/lavaan/index.html}}, and semopy\footnote{\url{https://semopy.com/}}),
and editing and analyzing DAGs (i.e., DAGitty\footnote{\url{http://www.dagitty.net/}}).
We believe this list of software solutions may be of interest to AI practitioners in helping them save valuable time when choosing the right tool to automate causal tasks.

The most popular choice is an open-source license type, and this reflects the great interest in sharing code and information across the AI research community.
The first benefit of that is flexibility. Researchers often need to access the source code of software implementations to eventually customize its functionalities according to a desired (yet not implemented) purpose. This would be highly unfeasible with closed and commercial software.
Another advantage of having open-source implementations is software security. According to Linus's law, "given enough eyeballs, all bugs are shallow" \citep{raymond1999cathedral}. That is, when all the source code for a project is made open to professionals worldwide, it is more likely that security checks could discover eventual flaws.

Furthermore, Table~\ref{tab:software_tools_qualities} shows that the CLI is the preferred frontend interface across such solutions. This aspect also reflects the AI research community viewpoint. Opting for CLI over the GUI brings some advantages, such as faster and more efficient computing, easier handling of repetitive tasks, lighter memory usage, and availability of the history of commands.
On the other hand, using CLI involves a steeper learning curve associated with memorizing commands and complex arguments, together with the need for correct syntax. This may explain why GUI is preferred in cases where the end-user does not have a programming background. Typical examples of that include physicians in healthcare facilities or product managers in finance companies, who prefer, in general, a more user-friendly product.

\begin{table}
\footnotesize 
\caption{Software tools within the cohort of papers useful to automate causal tasks. BSD: Berkeley Software Distribution, CD: causal discovery, CLI: common line interface, GPL: General Public License, GUI: graphical user interface.}
\label{tab:software_tools_qualities}
\begin{tabular}{>{\RaggedRight}p{5.5cm}|>{\RaggedRight}p{1.55cm}|>{\RaggedRight}p{4cm}|>{\RaggedRight}p{1.5cm}|>{\RaggedRight}p{2cm}}
\textbf{Name} & \textbf{License}\newline \textbf{type} & \textbf{Release}\newline \textbf{paper} & \textbf{Frontend}\newline \textbf{interface} & \textbf{Main\newline purpose}\\
\hline

\textit{bnlearn} & Open-source\newline (GPL) & \citet{scutari2010learning} & CLI (R) & BNs for CD\\
\hline
\textit{CaMML} by Bayesian Intelligence Pty Ltd & Open-source\newline (BSD) & n.a. & CLI (Bash)\newline and GUI & BNs for CD\\
\hline
\textit{CausalNex} by QuantumBlack, AI by McKinsey & Open-source\newline (Apache 2.0) & n.a. & CLI (Python) & BNs for CD\\
\hline
\textit{CompareCausalNetworks} & Open-source\newline (GPL) & \citet{heinze2018causal} & CLI (R) & BNs for CD\\
\hline
\textit{DAGgity} & Open-source\newline (GPL) & \citet{textor2016robust} & CLI (R) and GUI & Create and analyze causal diagrams\\
\hline
\textit{IBM SPSS Amos} by IBM Corp. & Commercial & n.a. & GUI & Create and analyze SCMs\\
\hline
\textit{lavaan} & Open-source\newline (GPL) & \citet{rosseel2012lavaan} & CLI (R) & Create and analyze SCMs\\
\hline
 \textit{PySMILE} by BayesFusion LLC & Commercial & n.a. & CLI (Python) & BNs for CD\\
\hline
\textit{Python Causal Discovery Toolbox} by Fentech & Open-source\newline (MIT) & \textit{\citet{kalainathan2020causal}} & CLI (Python) & BNs for CD\\
\hline
\textit{semopy} & Open-source\newline (MIT) & \citet{semopy}\newline \citet{meshcheryakov2021semopy} & CLI (Python) & Create and analyze SCMs\\
\hline
\textit{Tetrad} & Open-source\newline (GPL) & \citet{ramsey2018tetrad} & GUI & BNs for CD\\
 \hline
\end{tabular}
\normalsize
\end{table}

\section{Conclusion}
The concepts of causation and explanation have always been part of human nature, from influencing the philosophy of science to impacting the data mining process for knowledge discovery of today's AI.
In this study, we investigated the relationship between causality and XAI, by exploring the literature from both theoretical and methodological viewpoints, to reveal whether a dependent relationship between the two research fields exists. We provided a unified view of the two fields by highlighting which methodologies could be adopted to approach the bridge between these two fields and uncovering possible limitations.
As a result of the analysis, we found and formalized three main perspectives.

The \textit{Critics to XAI under the causality lens} perspective analyses how the lack of causality is one of the major limitations of current (X)AI approaches as well as the "optimal" forms to provide explanations.
Regarding the former, traditional AI systems are only able to detect correlation instead of true causation, which affects the robustness of models against adversarial attacks and of the produced explanations. This is of concern since pure associations are not enough to accurately describe causal effects.
Regarding the latter, optimal explanations may be characterized by being expressed according to the explainee's knowledge and domain terminology and being able to explain many effects with few causes. However, it is debated whether causal explanations (i.e., causal inference chains to a prediction) are the only useful ones in the XAI landscape. 
This first perspective states the problem and serves as a watch out.

The \textit{XAI for causality} perspective openly claims that XAI may be a basis for further causal inquiry. 
Despite the recognized limits of XAI explanations, they may be pragmatically thought of as starting points to generate hypotheses about possible causal relationships that scientists could then confirm. That is, XAI can only foster scientific exploration, rather than scientific explanation.
Although underrepresented in the final cohort, this perspective suggests a really thoughtful idea in our opinion.

The \textit{Causality for XAI} perspective supports the idea that causality is propaedeutic to XAI. 
This is realized in three manners.
First, some causal concepts (i.e., SCM and \textit{do}-operator) are leveraged to revisit existing XAI methods to empower them with causal inference properties. 
Second, the formal causal definition of CF (Sec.~\ref{sec:intro_rationale_causality}) is invoked to generate causal-CFEs using the SCM tool, which may also enable recourse. 
Third, and lastly, it is argued that, when a model is built on a causal structure, it is inherently an interpretable model. In a related way, making the inner workings of a causal model directly observable (e.g., through a DAG) makes the model inherently interpretable.

Among the three main perspectives, we believe \textit{Causality for XAI} to be the most promising one. Naturally, it comes with limitations. Much work in causal modeling is based on specific and (by far) non-unique causal views of the problems at hand. Interventions and CF make sense as long as the specified causal graph makes sense, which may hinder the generalization of their results. Overall, their causal claims depend on strong and often non-testable assumptions about the underlying data-generating process. 
On the other hand, however, this may be in line with what already happens in our life, and we should not request from AI more than we request from human beings. 
Another weak point is the interpretability of a causal model with hundreds of variables. In this scenario, a DAG would encode too much information and the complexity of the underlying SCM would rise exponentially with the number of modeled variables. This, however, is common to other simpler and more traditional approaches such as Decision Trees with hundreds of nodes.

We acknowledge three main limitations that may have led us to miss publications that could have potentially been included in the review: (i) the exclusion of non-peer-reviewed e-prints, (ii) the usage of only four databases, and (iii) not having extracted any references from the collected papers to enrich our search. The latter was motivated by the fact that, this being an unexplored field, the papers we collected were sufficient and significant enough to produce a first scenario. Obviously, as with any human-made assignment, the search process for relevant material may have been affected by the cognitive bias of the authors, who have brought their knowledge and assumptions in the study. 

We believe our results could be useful to a wide spectrum of readers, from upper-level undergraduate students to research managers in the industry, and have implications for practice, policy, and future research. Indeed, having a clear view of how the two concepts of causality and XAI are related can benefit both areas individually, as well as the joint research field. Considering our conceptual framework, future publications may be framed in a precise and rigorous way and have the potential to expand (or generate new flavors of) one of the identified perspectives.

All in all, our work disclosed how causality and XAI may be related in a profound way.
In our opinion, the \textit{Causality for XAI} perspective has great potential to produce significant scientific results and we expect the field to flourish the most soon.

\section{Funding Information}
This work was partially funded by: the European Union’s Horizon 2020 research and innovation programme under grant agreement No 952159 (ProCAncer-I), and the Regional Projects PAR FAS Tuscany - PRAMA and NAVIGATOR. The funders had no role in the design of the study, collection, analysis and interpretation of data, or writing of the manuscript.

\appendix
\section{Background notions}
\label{sec:appendix_background}
\subsection{Directed acyclic graphs}
\label{sec:appendix_background_dag}
\begin{figure}
    \centering
    \includegraphics[width=0.5\textwidth]{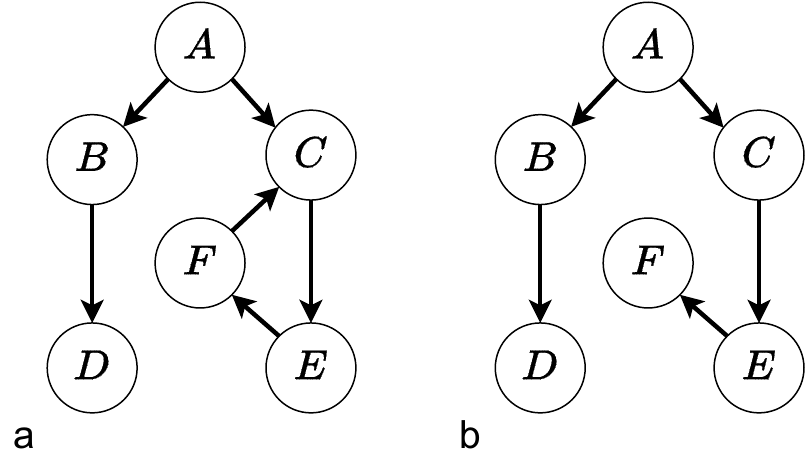}
    \caption{Examples of directed graphs: (a) directed cyclic graph, (b) directed acyclic graph (DAG).}
    \label{fig:simple_dag}
\end{figure}

From graph theory, a \textit{graph} consists of a set $V$ of vertices (i.e., variables) and a set $E$ of edges (i.e., relationships) that connect some pairs of vertices. A graph is \textit{directed} when all the edges are directed (i.e., marked by a single arrowhead).
In a directed graph, an edge goes from a \textit{parent} node to a \textit{child} node.
A \textit{path} in a directed graph is a sequence of edges such that the ending node of each edge is the starting node of the next edge in the sequence (e.g., nodes $A$, $B$, $D$ in Fig.~\ref{fig:simple_dag}). A \textit{cycle} is a path in which the starting node of its first edge equals the ending node of its last edge (e.g., nodes $C$, $E$, $F$ in Fig.~\ref{fig:simple_dag}a), and this represents mutual causation or feedback processes.
When a directed graph does not include directed cycles, it is called a \textit{directed acyclic graph} (DAG), and much of the discussion of causality and qualitative modeling is occupied by it \citep{pearl2009causality}.

\subsection{Bayesian networks}
\label{sec:appendix_background_bn}
A Bayesian network (BN) is a probabilistic graphical model that consists of two parts, a qualitative one based on a DAG, representing a set of variables and their dependencies, and a quantitative one based on local probability distributions for specifying the probabilistic relationships \citep{pearl1985bayesian}.
Let $\textbf{X}=[X_1, X_2, \dots, X_m]$ be a data matrix  with $n$ samples and $m$ variables. In the DAG $G=(V, E)$ of a BN, each node $V_k \in V$ represents the random variable $X_k$ in $\textbf{X}$, $k \in \{1,2, \dots, m\}$, and each edge $e \in E$ describes the conditional dependency between pairs of variables. The absence of an edge implies the existence of conditional independence.

The structure of the DAG can be constructed either manually, with expert knowledge of the underlying domain (knowledge representation), or automatically learned from a large dataset. 
In this regard, causal discovery (CD) denotes a broad set of methods aiming at retrieving the topology of the causal structure governing the data-generating process, using the data generated by this process. CD algorithms are commonly divided into two families: \textit{constraint}-based and \textit{score}-based.

\textit{Constraint}-based methods begin with fully-connected edges between random variables and leverage conditional independence tests to identify a set of edge constraints for the graph. By deleting relations if there is no statistical significance between variables, they narrow down the candidate graphs that explain the data and then try to determine the direction of the found relationships. Popular examples include the PC algorithm \citep{spirtes1991algorithm}, assuming no latent confounders (i.e., variables that are not directly observed but interact with the observables), and the Fast Causal Inference (FCI) algorithm \citep{spirtes2000causation}, whose results are asymptotically correct even in the presence of (possibly unknown) confounders.
Although constraint-based methods can handle various types of data distributions and causal relations, they do not necessarily provide complete causal information, since they output a set of causal structures satisfying the same conditional independence.
 
On the other hand, \textit{score}-based methods iteratively generate candidate graphs, assign them a relevance score to evaluate how well each one explains the data (i.e., “model fit”), and select the best one. Since enumerating (and scoring) every possible graph among the given variables is computationally expensive, these algorithms apply greedy heuristics to restrict the number of candidates.
Among them, Greedy Equivalence Search (GES) \citep{chickering2002optimal} is a well-known two-phase procedure that directly searches over the space of equivalence classes. Starting with an empty graph, at each step, it adds currently needed edges (if that increases fit), and then eliminates unnecessary edges in a pattern.

Regarding the quantitative part of which a BN consists, the local probability distributions can be either \textit{marginal}, for nodes without parents (root nodes), or \textit{conditional}, for nodes with parents. In the latter case, the dependencies are quantified by Conditional Probability Tables (CPTs) for each node given its parents in the graph. These quantities can be estimated from data in a process known as Parameter Estimation, two popular examples of which are the Maximum Likelihood approach and the Bayesian approach.

\begin{figure}
    \centering
    \includegraphics[width=0.75\textwidth]{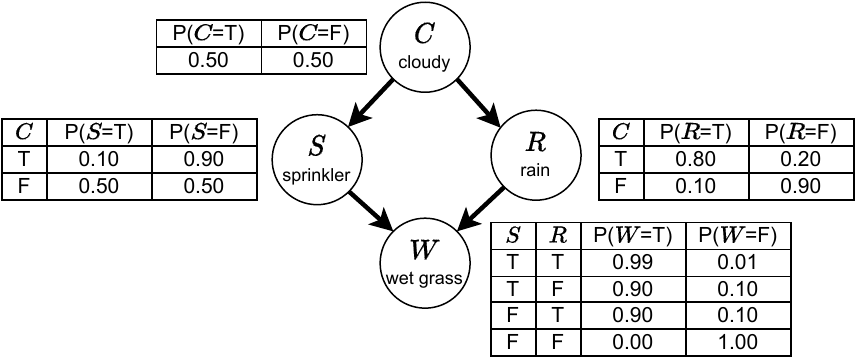}
    \caption{Example of a fully specified BN which models the probability of observing wet grass. In this (simplified) real-world scenario, grass can be wet either by turning on a sprinkler or by rainfall, and both can be influenced by the presence of clouds in the sky.}
    \label{fig:simple_BN}
\end{figure}

Once the DAG and CPTs are determined, a BN is fully specified and compactly represents the Joint Probability Distribution (JPD). An example of a fully specified BN is shown in Fig.~\ref{fig:simple_BN}. 
According to the \textit{Markov condition}, each node is conditionally independent of its non-descendants, given its parents. As a result, the JPD can be expressed in a product form:
\begin{equation}
    \label{eq:conditional_prob}
p(X_1, X_2, \dots, X_m) = \prod_{k=1}^{m} p(X_k | \mathbb{X}_{pa(k)})
\end{equation}
Where $\mathbb{X}_{pa(k)}$ is the set of parent nodes of $X_k$ and $p(X_k | \mathbb{X}_{pa(k)})$ is the conditional probability of $X_k$ given $\mathbb{X}_{pa(k)}$.
Thus, such a BN can be used for predictions and inference, that is, computing the posterior probabilities of any subset of variables given evidence about any other subset.

\subsection{Structural Causal Models}
\label{sec:appendix_background_scm}
Consider the set $\textbf{X}$ of variables associated with the vertices of a DAG. When each of them appears on the left-hand side (i.e., the dependent variable) of an equation of the type:  
\begin{equation}
\label{eq:scm}
X_k = f_k(\mathbb{X}_{pa(k)}, U_k), \quad k = 1, \dots, m
\end{equation}
that represents an autonomous mechanism, then the model is called a \textit{structural causal model} (SCM) \citep{pearl2009causality, scholkopf2021toward}.
In this equation, $f_k$ represents a deterministic function depending on the $X_k$'s parents in the graph (i.e., $\mathbb{X}_{pa(k)}$), and on $U_k$, which represents the exogenous variables (i.e., errors or noises due to omitted factors).
These noises are assumed to be jointly independent, and hence ensure that each structural equation can represent a general conditional distribution $p(X_k | \mathbb{X}_{pa(k)})$,
Recursively applying Eq.~\ref{eq:scm}, when the distributions of $U = \{U_1,\dots,U_m\}$ are specified, allows the computation of the entailed observational joint distribution $p(X_1, X_2,\dots,X_m)$, which, in turn, can be canonically factorized into Eq.~\ref{eq:conditional_prob}.
The advantages of using the SCM language include modeling unobserved variables (i.e., latent variables and counfounders), easily formalizing interventions, and computing CF.
Interventions and CF are defined through a mathematical concept called \textit{do}-operator, which simulates physical interventions by modifying a subset of structural equations (e.g., replacing them with a constant), while keeping the rest of the model unchanged.
Specifically, to compute the probability of CF, Pearl proposes a three-step procedure. Given a known SCM $M$ over the set $\textbf{X}$ of variables, let $x_{factual} = [X_1=x_1,X_2=x_2,\dots,X_m=x_m]$ be the evidence. To compute the probability of a counterfactual instance $x_{counterfactual}$, one needs to:
\begin{enumerate}
    \item \textit{abduction}: infer the values of exogenous variables in $U$ for $x_{factual}$, i.e., calculate $P(U|x_{factual})$;
    \item \textit{action}: intervene on $X=x_{factual}$ by replacing (some of) the equations by the equations $X=x_{counterfactual}$, where $x_{counterfactual} = [X_1=x_1',X_2=x_2',\dots,X_m=x_m']$, and thus obtain a new SCM $M'$;
    \item \textit{prediction}: use $M'$ to compute the probability of $P(x_{counterfactual}|x_{factual})$.
\end{enumerate}

\section{Study selection process}
\label{sec:appendix_studyselectionprocess}
Although we did not apply any temporal constraint to the search, we adopted some exclusion criteria in the process. We excluded works that were not written in English, articles from electronic preprint archives (e.g., ArXiv\footnote{\url{https://arxiv.org}}), book chapters, and theses.
In addition, we excluded too-short papers and/or papers of poor quality that hindered our ability to extract data meaningfully. We also deemed off-topic those papers that considered causality in the common and everyday sense of the term, not based on theoretical definitions. Indeed, they frequently present few occurrences of the causal domain terms, which were often either poorly contextualized or only present in the abstract/keywords of the article.

Regarding information sources, we selected Scopus, IEEE, Wos, and ACM because they cover a comprehensive range of AI works and provide powerful interfaces for retrieving the required data with limited restrictions.
Conversely, we excluded Google Scholar\footnote{\url{https://scholar.google.com/}}, SpringerLink\footnote{\url{https://link.springer.com/}}, and Nature\footnote{\url{https://www.nature.com/siteindex}} since they do not allow to formulate the query string with the same level of detail as the selected databases do, and, on the other hand, we excluded PubMed\footnote{\url{https://pubmed.ncbi.nlm.nih.gov/}}, since it provides this capability, but its coverage is restricted solely to the medical field.

As for the search strategy on the specified databases, the use of the wildcard made word-matching easier. For instance, \textbf{causal$*$} matched terms like \textit{causal} and \textit{causality}, while \textbf{expla$*$} matched terms such as \textit{explanation(s)}, \textit{explainable}, \textit{explainability}, \textit{explaining}, and \textit{explained}. 

On July 14, 2022, we utilized the research query on the four databases for the first time. We collected the retrieved publications and started analyzing them. Then, on September 5, 2022, we repeated the search in the same settings. This allowed us to refine our cohort of papers with new works that have been published in the meanwhile, therefore enriching our analyses. In general, although we utilized the same research query across the four databases (Sec.~ \ref{sec:methods_eligibility}), the actual query string was edited according to the specific syntax of each of them. In this regard, those strings are shown in Tab.~\ref{tab:query_strings}.

\begin{table}[t]
\caption{Query strings used for each database. AB, ABS: abstract; AK, KEY: keywords; TI: title.}
\label{tab:query_strings}
\centering
\resizebox{\textwidth}{!}{%
\begin{tabular}{p{2.2cm}|p{15cm}}
\hline
Database & Query string\\
\hline

Scopus & \texttt{TITLE-ABS-KEY(causal*) \textbf{AND}\newline
TITLE-ABS-KEY(expla*) \textbf{AND}\newline
TITLE-ABS-KEY(xai OR "explainable artificial intelligence" OR "explainable ai") \textbf{AND}\newline
TITLE-ABS-KEY("machine learning" OR ai OR "artificial intelligence" OR "deep learning")}\\
\hline

Web of\newline Science & \texttt{(TI=causal* OR AB=causal* OR AK=causal*)
\textbf{AND}\newline
(TI=expla* OR AB=expla* OR AK=expla*)
\textbf{AND}\newline
(TI=(xai OR "explainable artificial intelligence" OR "explainable ai") OR AB=(xai OR "explainable artificial intelligence" OR "explainable ai") OR AK=(xai OR "explainable artificial intelligence" OR "explainable ai"))
\textbf{AND}\newline
(TI=("machine learning" OR ai OR "artificial intelligence" OR "deep learning") OR AB=("machine learning" OR ai OR "artificial intelligence" OR "deep learning") OR AK=("machine learning" OR ai OR "artificial intelligence" OR "deep learning"))}\\
\hline

IEEE Xplore & \texttt{("Document Title":causal* OR "Abstract":causal* OR "Author Keywords":causal*) 
\textbf{AND}\newline
("Document Title":expla* OR "Abstract":expla* OR "Author Keywords":expla*) 
\textbf{AND}\newline
("Document Title":xai OR "Document Title":"explainable artificial intelligence" OR "Document Title":"explainable ai" OR "Abstract":xai OR "Abstract":"explainable artificial intelligence" OR "Abstract":"explainable ai" OR "Author Keywords":xai OR "Author Keywords":"explainable artificial intelligence" OR "Author Keywords":"explainable ai") 
\textbf{AND}\newline
("Document Title":"machine learning" OR "Document Title":ai OR "Document Title":"artificial intelligence" OR "Document Title":"deep learning" OR "Abstract":"machine learning" OR "Abstract":ai OR "Abstract":"artificial intelligence" OR "Abstract":"deep learning" OR "Author Keywords":"machine learning" OR "Author Keywords":ai OR "Author Keywords":"artificial intelligence" OR "Author Keywords":"deep learning")}
\\
\hline

ACM & \texttt{(Title:causal* OR Abstract:causal* OR Keyword:causal*)
\textbf{AND}\newline
(Title:expla* OR Abstract:expla* OR Keyword:expla*)
\textbf{AND}\newline
(Title:xai OR Title:"explainable artificial intelligence" OR Title:"explainable ai" OR Abstract:xai OR Abstract:"explainable artificial intelligence" OR Abstract:"explainable ai" OR Keyword:xai OR Keyword:"explainable artificial intelligence" OR Keyword:"explainable ai")
\textbf{AND}\newline
(Title:"machine learning" OR Title:ai OR Title:"artificial intelligence" OR Title:"deep learning" OR Abstract:"machine learning" OR Abstract:ai OR Abstract:"artificial intelligence" OR Abstract:"deep learning" OR Keyword:"machine learning" OR Keyword:ai OR Keyword:"artificial intelligence" OR Keyword: "deep learning")}
\\
\hline
\end{tabular}
}
\end{table}

\begin{figure}[t]
    \centering
    \includegraphics[width=0.5\textwidth]{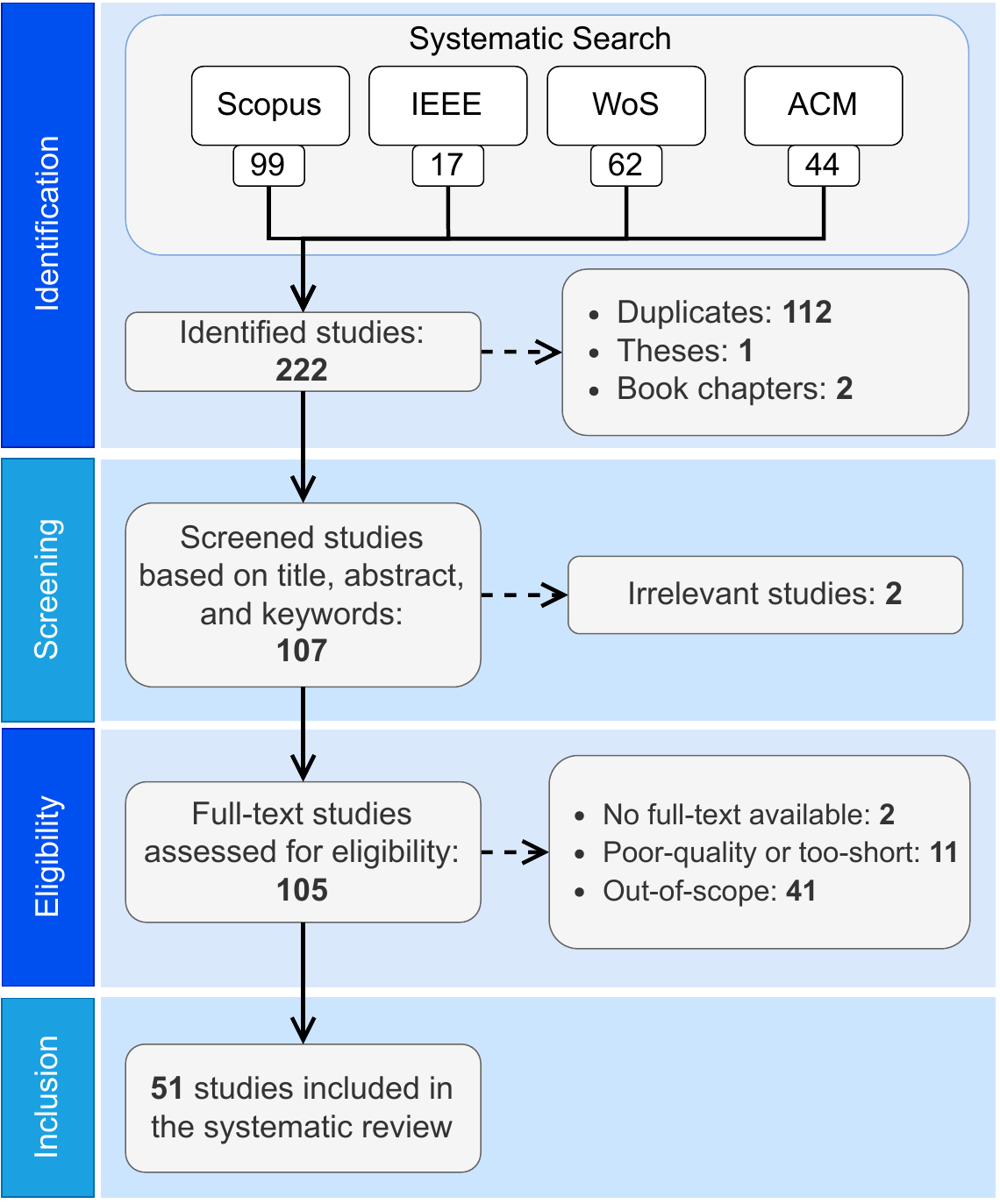}
    \caption{Flowchart of the study collection process, from identification, through screening, to eligibility and inclusion.}
    \label{fig:prisma_flowchart}
\end{figure}

Fig.~\ref{fig:prisma_flowchart} shows the process of identification, screening, eligibility, and inclusion of articles in our work.

From the search, we obtained the following number of records from the four databases: $99$ (Scopus), $17$ (IEEE), $62$ (WoS), and $44$ (ACM). As a result, we collected a total of $222$ publications.
Upon extraction of query results from the databases, we operated the identification phase. For the retrieved records, we extracted the BibTeX files and uploaded them into a popular reference manager application by Elsevier, namely Mendeley\footnote{\url{https://www.mendeley.com/}}, desktop version 1.19.8. We then utilized its \textit{Check for Duplicates} feature to perform duplicate removal. Then, we removed one thesis and two book chapters, according to the defined exclusion criteria. After these steps, the joint output was $107$ publications.

During the screening phase, we examined independently the resulting works by title, abstract, and keywords to verify and ensure that proper results were retrieved by the query.
Whenever both authors deemed a paper irrelevant, it was discarded from the cohort. Specifically, two publications were hereby discarded. Instead, publications for which the authors agreed on the inclusion, together with those on which they disagreed, passed to the next phase.

Next, in the eligibility phase, we first checked for the availability of full-text manuscripts for the records in the cohort. We excluded two studies as we could not access their full text. We then jointly analyzed the available full-text publications to remove papers that were clearly out of scope, together with poor-quality or too-short papers. As a result, we identified $11$ poor-quality or too-short papers and $41$ out-of-scope works. 
Lastly, once we reached a common decision for each of the publications, we collected the final cohort of studies to be included in the review.

\newpage
\bibliographystyle{elsarticleHarv} 
% \bibliography{casRefs}
\bibliography{main}

\end{document}